\documentclass{article}

\usepackage{PRIMEarxiv}
\usepackage{natbib}
\usepackage[utf8]{inputenc} 
\usepackage[T1]{fontenc}    
\usepackage{hyperref}       
\usepackage{url}            
\usepackage{booktabs}       
\usepackage{amsfonts}       
\usepackage{nicefrac}       
\usepackage{microtype}      
\usepackage{lipsum}
\usepackage{fancyhdr}       
\usepackage{graphicx}       
\graphicspath{{media/}}     

\usepackage{subcaption}
\usepackage{multirow, makecell}
\usepackage{array}
\newcolumntype{P}[1]{>{\centering\arraybackslash}p{#1}}
\usepackage{amssymb}
\usepackage{amsthm}
\DeclareMathSymbol{\mh}{\mathord}{operators}{`\-}
\usepackage{amsmath}
\DeclareMathOperator*{\argmin}{arg\,min}
\DeclareMathOperator{\EX}{\mathbb{E}}

\title{Hybrid Machine Learning Modeling of Engineering Systems - A Probabilistic Perspective Tested on a Multiphase Flow Modeling Case Study
}

\author{
	Timur Bikmukhametov \\
	Dept. of Chemical Engineering \\
	Norwegian University of Science and Technology (NTNU) \\
	NO-7491 Trondheim \\
	\texttt{timur.bikmukhametov@gmail.com} \\
	\And
	Johannes J{\"a}schke \\
	Dept. of Chemical Engineering \\
	Norwegian University of Science and Technology (NTNU) \\
	NO-7491 Trondheim \\
	\texttt{johannes.jaschke@ntnu.no} \\
}

\begin{document}
\maketitle

\begin{abstract}
	
To operate process engineering systems in a safe and reliable manner, predictive models are often used in decision making. In many cases, these are mechanistic first principles models which aim to accurately describe the process. In practice, the parameters of these models need to be tuned to the process conditions at hand. If the conditions change, which is common in practice, the model becomes inaccurate and needs to be re-tuned. In this paper, we propose a hybrid modeling machine learning framework which allows to tune first principles models to process conditions using two different types of Bayesian Neural Networks. Our approach not only estimates the expected values of the first principles model parameters but also quantifies the uncertainty of these estimates. Such an approach of hybrid machine learning modeling is not yet well described in the literature, so we believe this paper will provide an additional angle at which hybrid machine learning modeling of physical systems can be considered. As an example, we choose a multiphase pipe flow process for which we constructed a three-phase steady state model based on the drift-flux approach which can be used for modeling of pipe and well flow behavior in oil and gas production systems with or without the neural network tuning. In the simulation results, we show how uncertainty estimates of the resulting hybrid models can be used to make better operation decisions.

\end{abstract}



\section{Introduction}
\label{Introduction}
Today, process operation decisions are increasingly made on the basis of process models. Traditionally, process models are derived from first principles such as mass, momentum and energy conservation laws. This approach is called first principles or mechanistic modeling (\cite{pantelides2013online}). The main advantage of these models is that they are built based on knowledge about the system and are understandable to the knowledgeable user. A conceptually different approach is based on using process data to learn a model of the process (\cite{rasheed2019digital}). Data-driven approaches of this kind become increasingly popular due to the progress in data handling, analytics and machine learning techniques.

In the recent years, mechanistic models of process plants and equipment together with machine learning and virtual reality technologies have formed the Digital Twin framework which may be used for monitoring of process conditions, equipment health and plant efficiency, and support critical decisions on plant operation quickly and effectively with minimal human intervention (\cite{rasheed2019digital}). Digital Twins offer significant potential in operating cost reduction, however, they require in-time automatic or semi-automatic model recalibration to be successfully used in industrial systems (\cite{rasheed2019digital}). 

In all but the most trivial cases, first principles model parameters need to be tuned, such that the model predictions accurately match the plant. Such tuning can be conducted in experimental laboratories, onsite at a process plant or, in many cases, both options are required (\cite{pantelides2013online}). During operation, the predictive accuracy of the obtained models will depend on several factors. One factor is the measurement accuracy. Process measurements may drift over time, as such the model calculates its outputs based on wrong inputs, and the model accuracy drifts accordingly (\cite{bikmukhametov2018statistical}). Another important factor is that process conditions such as material or fluid properties may change over time and differ from the tuning conditions which may lead to inaccurate predictions and economic loss (\cite{mazzour2008measurement}, \cite{amjad2003closed}). In addition, equipment in process plants often degrades over time which makes the predictive model of the equipment behavior inaccurate (\cite{gorjian2010review}). As such, it is very important to identify under which conditions the tuned mechanistic model becomes unreliable to use and needs recalibration, and this is the topic of the present paper.

One promising method to estimate the need for model recalibration and understanding its predictive capabilities is to quantify uncertainty and bias of model predictions. If the prediction uncertainty and/or bias is high, then the model recalibration or restructuring is required, because the produced estimates cannot be trusted.

In the literature, there are different contributions which report methods to quantify uncertainty of first principles models. One approach relies on checking sensitivities of model predictions with respect to the change in model parameters, typically referred as Sensitivity Analysis (\cite{ratto2007uncertainty}, \cite{digiano2004uncertainty}). In this approach, selected parameters are perturbed, the change of the model output is recorded and then the ratio of the output change to the parameter change is computed. Although easy to implement, this method does not consider the uncertainty of the data and model. Therefore, it estimates the sensitivity of the model with respect to the input parameters. The main result from such sensitivity analysis is the conclusion about the parameters to which the fitted mechanistic model is the most sensitive to, but not in general about how well the model matches the process. 

A more informative approach about model uncertainty is based on the Bayesian framework. In this approach, the uncertainty of parameters is estimated based on the tuned model to the data using various methods. One approach is to use Markov Chain Monte Carlo (MCMC) method which computes the probability of the parameters via a posterior parameter distribution over the proposed parameters' priors (\cite{aldebert2018community}, \cite{leil2014bayesian}). MCMC is a powerful concept of estimating unbiased posterior distributions, however, it may be very slow for many practical applications and accelerating MCMC methods is an active field of research (\cite{robert2018accelerating}). When applied to parametric uncertainty estimation of mechanistic first principles models, this can be an issue, if, for instance, the mechanistic model needs a time consuming iterative procedure to be solved. \cite{shrestha2009novel} addressed this problem by approximating MCMC results using neural networks, however, in this case, MCMC simulations still need to be performed.

In this paper, we propose to tune the parameters of the first principles model using Bayesian Neural Networks (BNNs). This is done by, first, learning the distribution of the neural network parameter values from the process data and then propagating this distribution through the first principles model to give a distribution of the model output. This is then used to quantify the first principles model uncertainty. As such, the main idea is to utilize uncertainty quantification capabilities of Bayesian Neural Networks when estimating first principles model parameters.

Previously in the literature, authors have adapted maximum likelihood neural networks for model parameter tuning, see the works by   \cite{psichogios1992hybrid}, \cite{ahmadi2018comparison}, \cite{anifowose2017hybrid}, \cite{klyuchnikov2019data}, \cite{onwuchekwa2018application}, \cite{kanin2019predictive}. However, using the maximum likelihood approach, it is not possible to estimate the uncertainty of the first principles model parameters, while our approach allows doing so.

For the training of BNNs, we build upon work by \cite{blundell2015weight} who used variational approximations for Bayesian learning of neural networks and work by \cite{gal2015dropout} who introduced the concept of Markov Chain (MC) Dropout techniques for deep neural networks. We compare the performance of both approaches and give recommendations on the conditions for combining each type of Bayesian Neural Networks with first principles models.

As an example of the first principles model, we create a complex, space discretized model of a three-phase (oil, gas and water) flow in pipes which considers the slip effect and mass transfer between the gas and liquid phases. We show how the friction coefficient of this model can be tuned using a Bayesian Neural Network and how uncertainty of these estimates can be quantified using this approach. This approach is promising in terms of scalability and potentially can be applied to systems of a relatively big size.

As such, the main contribution of this work is that we provide and exemplify a new flexible framework for hybrid machine learning modeling with uncertainty estimation of complex mechanistic first principles models. In addition, we discuss how the obtained results can be interpreted for the purpose of model recalibration. 


The paper is organized as follows. Section~\ref{Concept} introduces the proposed concept of tuning first principles models parameters by means of Bayesian Neural Networks. Section~\ref{Adaptation} describes the multiphase flow model which is used as the example of the first principles model to be tuned by BNNs and shows adaptation of the concept described in Section~\ref{Concept} for this model. In Section~\ref{Results}, we provide the simulation results and discuss them. In Section~\ref{Conclusions}, we make conclusions from our work.

\section{Proposed concept of parameter estimation of first principles models using Bayesian Neural Networks}

\subsection{General description of the proposed concept}

\begin{figure}[!t]
	\begin{center}
		\includegraphics[width=1.0\columnwidth]{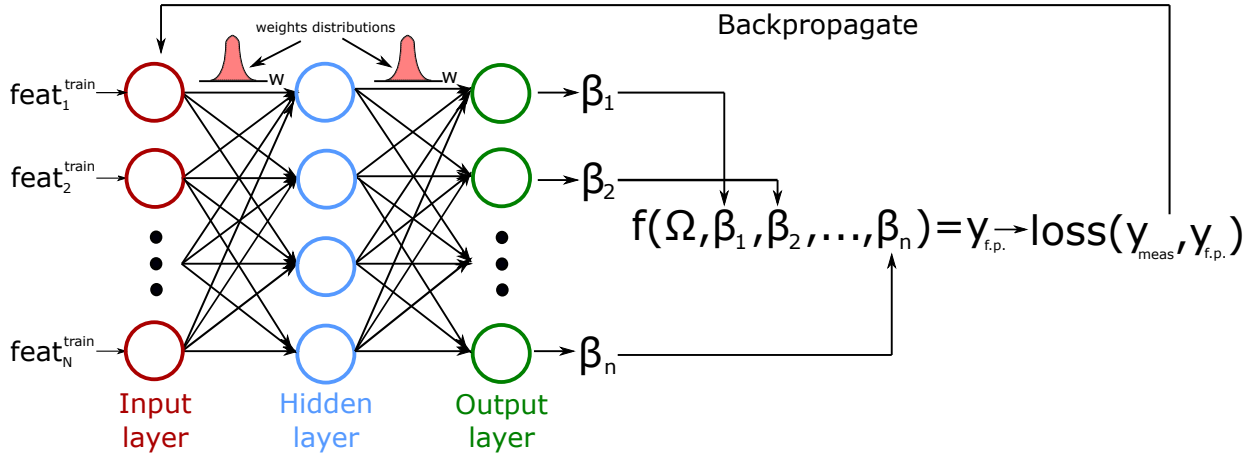}
		\caption{Training Bayesian Neural Networks for first principles models tuning} 
		\label{fig:training}
	\end{center}
\end{figure}

The main idea of the proposed framework is similar to any first principles model tuning to data - select parameters to be tuned or introduce coefficients which are adjusted to get an accurate fit to the data. A schematic representation of this process is shown in Figure~\ref{fig:training}. The main difference between the maximum likelihood tuning is that we learn the distribution of weights and biases represented by the approximate variational distributions instead of single point estimates. We do this by two BNN models that are described in Section~\ref{Bayes by Backprop} and ~\ref{MC Dropout_section} below.

At the prediction (test) stage, shown in Figure~\ref{fig:test}, the test features are fed into the trained Bayesian Neural Network. Then, a specified number of samples from the trained distribution of weights and biases is taken. These sampled parameters produce distributions of the tuned first principles model parameters. As a result, a distribution of the modeled output by a mechanistic first principles model is obtained. From the distribution, the mean and variance can be calculated. The mean value will correspond to the maximum a posterior estimate of the variable while the variance will correspond to the prediction uncertainty.

It is important to note that the proposed approach considers the uncertainty of model parameters based on the training data and does not account for the uncertainty of the model structure. As such, we assume that in order to apply this method, one should consider the model which is structurally correct such that it describes the general plant behavior well while may not consider all the complexities of the modeled phenomenon.

\begin{figure}[!t]
	\begin{center}
		\includegraphics[width=1.0\columnwidth]{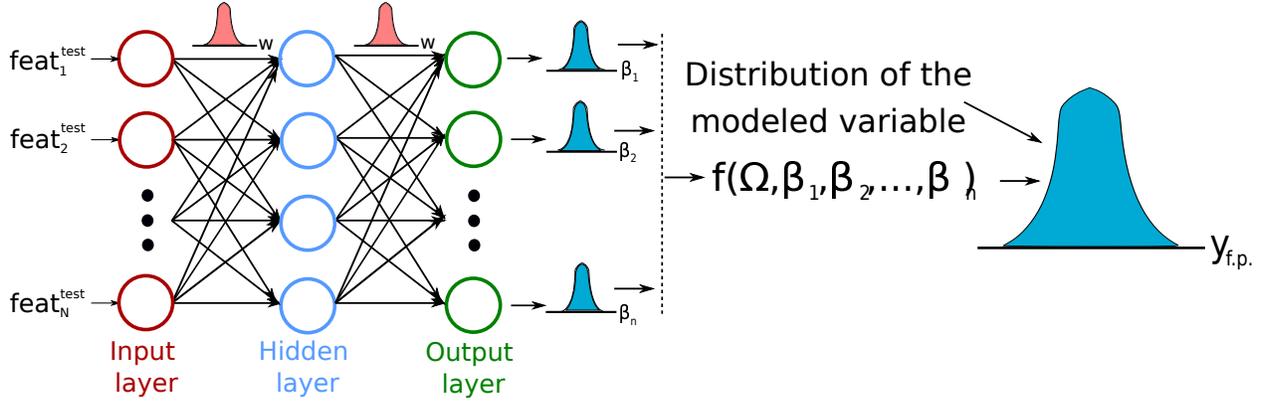}
		\caption{Prediction stage of Bayesian Neural Networks for predicting the mean and variance of the modeled variable} 
		\label{fig:test}
	\end{center}
\end{figure}

\label{Concept}
\subsection{General form of first principles models}
To describe the proposed hybrid machine learning framework of estimating parameters of first principles models in more detail, first, we define the general mathematical form of a steady state first principles model as:

\begin{equation}\label{eq:fp_model}
f(\Omega, B) = 0
\end{equation} 

where  $\Omega = [\omega_1, \omega_2, ..., \omega_n]^T$ denotes a vector of states of the process which is described by the model (e.g. pressure, temperature, medium properties, etc.) and $B = [\beta_1, \beta_2, ..., \beta_n]^T$ denotes a vector of the first principles model parameters. Also, user dependent and known input variables will typically be present in the model, but we do not include them explicitly to keep the notation simple.

Let us further denote $y_{meas}$ as a vector of process measurements. We assume that given a set of parameters $B$, a first principles model can produce an estimate of the process measurement $y_{meas}$ when solved with respect to it, i.e.:

\begin{equation}\label{eq:fp_model_solution}
 \hat{y}_{f.p.} = g(\Omega, B)
\end{equation}

where  $\hat{y}_{f.p.}$ is an estimate of $y_{meas}$ obtained by solving the first principles model $f(\Omega, B)$ for $\Omega$ and evaluating the output equation $g(\Omega, B)$.

In case when the mechanistic model describes the process accurately, $\hat{y} \simeq y_{meas}$. This can be the case when the model structure is correct and the model parameters $B$ are accurately tuned to the process data. In this work, we do not discuss the general framework of how to construct an accurate representation of the mechanistic first principles model. Instead, given a sequence of steady state process data $X = [(x^{(1)}, y^{(1)}), (x^{(2)}, y^{(1)}) ..., (x^{(n)}, y^{(n)})]$, we want to estimate the values of the parameters vector $B$ and the uncertainty of the predictions based on the data. 

\subsection{Bayesian Neural Networks}

\subsubsection{Bayesian Learning Framework}
Bayesian Neural Networks is a family of artificial neural networks whose weights are represented by distributions rather than by point estimates as in the maximum likelihood approach (\cite{neal2012bayesian}). These distributions represent our beliefs about the values of the parameters. Bayesian learning of neural networks (and any other machine learning models) starts with defining prior distributions $P(W)$ of neural network parameters, i.e. weights and biases. The prior distributions express our prior beliefs about the neural network parameters before we fit it to any data (\cite{neal2012bayesian}). In the learning process, the model parameters are updated according to the Bayes' rule:

\begin{equation}\label{eq:bayes_rule}
P(W|X) = \frac{P(X|W)P(W)}{P(X)}
\end{equation}

where $X$ denotes the vector of observed data and the associated target variable (in case of this paper - measurement) $[(x^{(1)}, y^{(1)}), (x^{(2)}, y^{(1)}) ..., (x^{(n)}, y^{(n)})]^T$ and $W$ denotes the vector of neural network parameters (weights and biases).

In Eq.~\ref{eq:bayes_rule}, $P(W|X)$ is the posterior distribution of the model parameters which is a result of the update of our prior beliefs about them represented by $P(W)$ after fitting the model to the data. $P(X|W)$ is called a likelihood function which represents neural network predictions for a given set of parameters (weights and biases). $P(X)$ is the normalizing constant which ensures that the probability sums to one. 

From the Bayes' rule we see that the learning process of model parameters is done in a natural way, such that we propose our beliefs about how the model should look like and then update it according to the observed data. Also, the main advantage of Bayesian learning is obtaining uncertainty of model parameters in addition to maximum likelihood (or maximum a posterior) point estimates. As the result, we are be able to estimate uncertainty of the target (modeled) variable during inference as the following:

\begin{equation}\label{eq:bayes_prediction}
\begin{gathered}
P(y^{(n+1)}|x^{(n+1)}, (x^{(1)}, y^{(1)}), ...,(x^{(n)}, y^{(n)})) \\ = \int P(y^{(n+1)}|x^{(n+1)}, W)P(W|(x^{(1)}, y^{(1)}), ...,(x^{(n)}, y^{(n)})) dW
\end{gathered}
\end{equation}
where $(x^{(1)}, y^{(1)}), (x^{(2)}, y^{(1)}) ..., (x^{(n)}, y^{(n)})$ denote the observed data points and $x^{(n+1)}$ denotes the point for which the target variable is estimated.

The main disadvantage of the Bayesian approach for neural networks learning is that the probability distribution $P(X)$ is high dimensional and analytically intractable (\cite{bishop1997bayesian}). For this reason, various approximations of posterior parameter distributions $P(X|W)$ are used. In this work, we use variational inference (\cite{blundell2015weight}) and Markov Chain (MC) Dropout (\cite{gal2015dropout}) approaches as approximations of the posterior distribution, which are described in the next sections in more detail. The main reason why these methods have been chosen is the fact that they can be used for big data sets in the context of process engineering systems, as such the methods proposed in this paper can be applied to many systems of interests. 

\subsubsection{Bayes by Backprop}
\label{Bayes by Backprop}
One popular approach to approximate the exact posterior distribution of neural network parameters $P(W|X)$ is to use a variational approximation of it, as proposed by \cite{hinton1993keeping} and further developed \cite{graves2011practical}. The main idea is to use a variational distribution $q(W|\theta)$ on the weights parameterized by $\theta$ and find such parameters $\theta$ of the approximate distribution $q(W)$ which minimize the Kullback-Leibler (KL) divergence (which is a measure of distributions similarity), with respect to the exact posterior distribution $P(W|X)$ (\cite{blundell2015weight}):

\begin{equation}\label{eq:KL_div}
\theta^{\star} = \underset{\theta} \argmin KL[q(W|\theta)||P(W|X)] = \argmin KL[q(W|\theta)||P(W)] - \EX_{q(W|\theta)}[logP(X|W)]
\end{equation}

From Eq.~\ref{eq:KL_div}, the resulting cost to be minimized is:

\begin{equation}\label{eq:KL_cost}
J(X, W, \theta) = KL[q(W|\theta)||P(W|X)] - \EX_{q(W|\theta)}[logP(X|W)]
\end{equation}
where cost function $J(X, W, \theta)$ is called a variational free energy (\cite{blundell2015weight}). 

\cite{blundell2015weight} showed that the cost $J(X, W, \theta)$ can be approximated as:

\begin{equation}\label{eq:KL_cost_approx}
J(X, W, \theta) \approx  \sum\limits_{i=1}^{n} q(w^{(i)}|\theta) - log P(w^{(i)}) - log P(X|w^{(i)})
\end{equation}

where $w^{(i)}$ denotes a weight sample from the variational posterior distribution $q(w^{(i)}|\theta)$.

This approach of training Bayesian Neural Networks is called the Bayes by Backprop (BBP) algorithm. At the inference stage of BBP, weight samples are drawn from the variational posterior distribution $q(W|\theta)$ which substitutes the exact posterior distribution $P(W|X)$ in Eq.~\ref{eq:bayes_prediction}. As a result, the mean estimate of the target variable as well as the uncertainty are estimated from the resulting approximated distribution $P(y^{(n+1)}|x^{(n+1)}, (x^{(1)}, y^{(1)}), ..., (x^{(n)}, y^{(n)}))$. 

\subsubsection{Markov Chain (MC) Dropout}
\label{MC Dropout_section}
The second approach to approximate the posterior distribution of the neural network parameters is Markov Chain Dropout. The theoretical foundation of Markov Chain Dropout is based on the fact that neural networks with applied dropout for each weight layer is mathematically equivalent to variational inference in the deep Gaussian Process. The derivation of the method is outside the scope of this paper and the interested reader is referred to the original article by \cite{gal2015dropout} and the appendix of the referred paper.

The outcome of \cite{gal2015dropout} derivations is that the exact posterior distribution of weights $P(W|X)$ can be approximated by an approximate variational distribution $q(W|\theta)$ via minimizing the following objective function:

\begin{equation}\label{eq:mc_dropout_loss}
J(X, W, p_{mc}) = - \frac{1}{N} \sum\limits_{i=1}^{N} log P(x^{(i)}|w^{(i)}) + \frac{1-p_{mc}}{2N} \| W\|^2
\end{equation}
where $p_{mc}$ denotes the MC Dropout probability, $w^{(i)}$ denotes the sample drawn from the variational distribution $q(W|\theta)$.

The advantage of this method is that the loss expressed in Eq.~\ref{eq:mc_dropout_loss} is the same as we would have had during  training of a traditional maximum likelihood neural network using dropout.

At the inference time, the dropout probability is kept, as such we obtain the mean and the variance of the posterior distribution $P(y^{(n+1)}|x^{(n+1)}, (x^{(1)}, y^{(1)}), ..., (x^{(n)}, y^{(n)})$ as the following:

\begin{equation}\label{eq:mean_mc_drop}
E(y) \approx \frac{1}{T} \sum_{t=1}^{T} P(x^{(i)}|w^{(i)})
\end{equation}

\begin{equation}\label{eq:mean_mc_var}
var(y) \approx \sigma^2 + \frac{1}{T} \sum_{t=1}^{T} P(x^{(i)}|w^{(i)}) P(x^{(i)}|w^{(i)}) - E(y)^TE(y)
\end{equation}

where $T$ denotes the number of stochastic passes through the neural network,$E(y)$ - the expected value of the target variable, $var(y)$ - the variance of the produced estimates of the target variable, $\sigma^2$ - the data noise (irreducible error).

The part of Eq.~\ref{eq:mean_mc_var} excluding $\sigma^2$ represent model (or epistemic) uncertainty which can be reduced if more data is collected and used for training, while $\sigma^2$ is the irreducible error and called aleatoric uncertainty.

\section{First principles model description and solution method.}

\label{Adaptation}
\subsection{System description.} We consider a problem of pressure drop tuning of multiphase flows in pipes. More specifically, we study a three-phase multiphase flow model which aims to replicate a steady state multiphase flow behavior in oil and gas production systems. This is a complex phenomenon and has been an important research topic since 1950's. The first attempt to model multiphase flow pressure drop in pipes was based on empirical correlations proposed by \cite{lockhart1949proposed}. Nowadays, the research and industrial standard of modeling this phenomenon is based on first principles such as mass, momentum and energy equations supported by some lab or field correlations to close the mathematical set of equations. The mechanistic approach allows a much more accurate description of complex flow phenomena including dynamic flow situations such as severe slugging (\cite{taitel1986stability}).

Due to the complexity of multiphase flow phenomena, it is difficult to create such closure laws and empirical relations at the lab which accurately hold at field conditions. We consider OLGA (\cite{bendiksen1991dynamic}) simulation results as the "true" field conditions. OLGA is the industrial standard of multiphase flow simulations, and we use it as a reference for data generation. The three-phase flow model developed in this work is used as the model which we would like to adjust to the "true" plant data generated by OLGA. Our first goal is to build a first principles model which produces relatively accurate results. However, it will not be able to exactly match the "true" plant data generated by OLGA due to different model formulation and different way of computing phase and mixture densities and viscosities which influence the friction loss values. To account for this inaccuracy between OLGA and our model, we use Bayesian Neural Networks which adjust the constructed model parameters and also estimate the uncertainty of the tuned predictions. 

\begin{figure}[!t]
	\begin{center}
		\includegraphics[width=1.0\columnwidth]{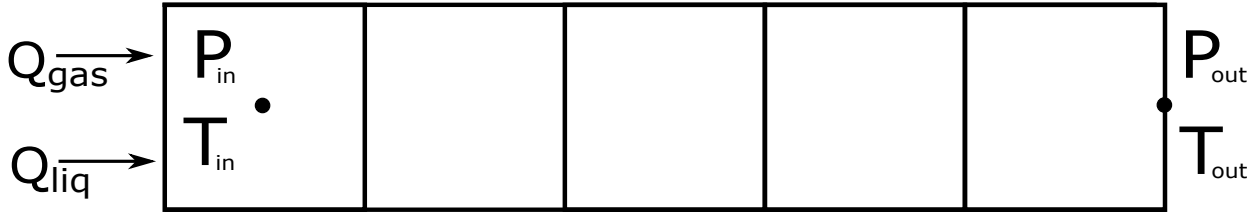}
		\caption{Straight pipeline plant setup used for case studies. The goal is to predict $P_{in}$ given the following boundary conditions: $Q_{gas}$, $Q_{liq}$, $T_{in}$, $P_{out}$, $T_{out}$. When $P_{in}$ is predicted, other parameters such as phase velocities or flow regimes can be identified.} 
		\label{fig:case_study}
	\end{center}
\end{figure}

We selected a relatively simple plant setup to keep the discussion of the simulated phenomena under control. We model a straight horizontal pipe flow (see Fig.~\ref{fig:case_study}), such that pressure drop in the pipe is caused only by friction. Table~\ref{properties} shows all the required boundary conditions which are constant in the training and test sets. We consider a relatively short pipeline of 1000 m with small pipe diameter of 0.2 m to get a high pressure drop using high flowrate values. This allows avoiding long simulations during the training and tuning process and larger error between OLGA simulations and our model but does not affect the application procedure of the proposed concept.

In this work, the inlet and outlet temperatures are assumed to be equal, so the fluid flow is assumed to be isothermal. The outlet pressure is required to get the solution of the drift-flux model using SIMPLE integration scheme (\cite{wang2016numerical}, \cite{spesivtsev2013comparison}) which we later discuss in the paper and the value for this pressure is kept at 10 bar. The fluid has a relatively low Gas-Oil-Ratio (GOR) and Water Cut (WC) values and typical values of standard gas, oil and water densities as well as bubble point pressure value.

\begin{table}[!t]
	\begin{center}
		\begin{tabular}{l|l}
			\centering
			Parameter & Value  \\ 
			\hline
			Pipe length & 1000 m \\
			Pipe diameter ($D_{pipe}$) & 0.2 m \\
			Hydraulic roughness ($\epsilon$) & 3$e^{-5}$ m \\
			Outlet pressure $(P_{out})$ & 10 bar \\
			Outlet temperature $(T_{out})$ & 25 $\,^\circ$C \\
			Inlet temperature $(T_{out})$ & 25 $\,^\circ$C \\
			Gas-oil ratio ($GOR$) & 50 $Sm^3/Sm^3$ \\
			Water cut ($WC$) & 0.3  \\ 	
			Bubble point pressure  $(P_{bp})$ & 50 bar \\
			Bubble point temperature $(T_{bp})$ & 20$\,^\circ$C \\ 
			Standard oil density $(\rho_{\bar{o}})$ & 867 $kg/m^3$\\
			Standard water density $(\rho_{\bar{w}})$ & 1020 $kg/m^3$ \\
			Standard gas density $(\rho_{\bar{g}})$ & 0.997 $kg/m^3$ \\
		\end{tabular}
	\end{center}
	\caption{Fluid properties, boundary conditions and pipe dimensions use for case studies in the training and test sets}
	\label{properties}
\end{table}

\subsection{The first principles model}
\label{Model}

In this section, we introduce the main part of the first principles model which is used as the example of the model to be tuned by Bayesian Neural Networks. 

\subsubsection{Fluid properties model}
In order to model multiphase fluid flow motion in pipes, first, a model of the fluid properties at different conditions have to be created. There are two main approaches to model fluid properties of a petroleum liquid: compositional model and Black Oil (BO) model. In the compositional formulation, mass or molar fractions of the petroleum components are specified and then Equations of State are solved to compute the required fluid properties (\cite{whitson2000phase}). In the Black Oil formulation, oil and gas are considered separately and their properties are computed using correlations which are based on several model properties such as solution gas-oil ratio, water cut, etc (\cite{whitson2000phase}). The compositional approach can be generally more accurate and should be considered when a fluid composition is available. However, this is typically not the case and often only Black Oil properties are measured in the lab for petroleum calculations. 

In this work, we use Standing Black Oil correlations (\cite{standing1947pressure}) for solution gas-oil ratio $R_{so}$ and bubble point pressure $P_{bp}$. Standing correlations are suitable for the conditions we use in our problem, but in practice, other correlations which are more suitable for the conditions at hand can also be used and this will not change neither the solving procedure of the first principles models nor the tuning part using Bayesian Neural Networks.

Since we would like to achieve the results which are relatively close to OLGA simulation outcomes, we use fluid properties correlations of the OLGA Black Oil model which are mainly based on the work by \cite{mccain1991reservoir}. The summary of the used correlations are shown in Table~\ref{BO_properties}.

\begin{table}
	\begin{center}
		\begin{tabular}{l|l}
			\centering
			Fluid property & Correlation  \\ 
			\hline 
			Critical pressure ($P_{crit}$) & \cite{mccain1991reservoir} \\ 
		
			Critical temperature ($T_{crit}$) & \cite{mccain1991reservoir} \\ 
		
			Oil formation volume factor ($B_o$) & \cite{mccain1991reservoir}  \\ 
		
			Water formation volume factor ($B_w$) & \cite{mccain1991reservoir}  \\ 
		
			Gas compressibility factor ($z$) & \cite{dranchuk1975calculation}  \\
			
			Dead oil viscosity ($\mu_{doil}$) & \cite{egbogah1990improved} \\ 
			
			Saturated oil viscosity ($\mu_{loil}$) & \cite{beggs1975estimating}  \\ 
			
			Gas viscosity ($\mu_{gas}$) & \cite{lee1966viscosity}  \\
			
			Water viscosity ($\mu_{water}$) & \cite{mccain1991reservoir}  \\
			
			Oil-gas surface tension ($\sigma_{o-g}$) & \cite{abdul2000estimation}  \\
		\end{tabular}
	\end{center}
\caption{Fluid properties correlations using within the Black Oil formulation}
\label{BO_properties}
\end{table}

Apart from that, we compute some of the properties based on the material balance. In particular, the formation liquid volume factor is computed as:

\begin{equation}\label{eq:b_l}
B_{l} = B_{o} \bigg(\frac{1}{1 + WOR} \bigg) + B_{w} \bigg(\frac{WOR}{1 + WOR} \bigg)
\end{equation}
where $B_{l}$ denotes the formation liquid volume factor and $WOR$ - the water-oil ratio which is computed as $\frac{WC}{1-WC}$ where $WC$ denotes the water cut.

Densities of oil, gas and water are also computed from the material balance: 

\begin{equation}\label{eq:densities}
\rho_{o,g,w} = \frac{\rho_{\bar{o},\bar{g},\bar{w}}}{B_{o,g,w}}
\end{equation}
where $\rho_{o,g,w}$ denotes the oil, gas and water density respectively at local conditions, $\rho_{\bar{o},\bar{g},\bar{w}}$ - the oil, gas and water density respectively at standard conditions, $B_{o,g,w}$ - the oil, gas and water formation volume factor respectively at local conditions.

Mixture density and viscosity are computed based on the volumetric basis:

\begin{equation}\label{eq:mix_density}
\rho_{mix} = \alpha_g \rho_{gas} +  (1-\alpha_g) \rho_{liq}
\end{equation}
where $\rho_{mix}$ denotes the mixture density and $\alpha_g$ - the gas volume fraction.

\begin{equation}\label{eq:mix_viscosity}
\mu_{mix} = \alpha_g \mu_{gas} +  (1-\alpha_g) \mu_{liq}
\end{equation}
where $\mu_{mix}$ denotes the mixture viscosity.

The solution gas-liquid ratio $R_{sl}$ is computed as:

\begin{equation}\label{eq:r_sl}
R_{sl} = R_{so} \bigg(  \frac{1}{1 + WOR} \bigg)
\end{equation}
where $R_{so}$ denotes the solution gas-oil ratio at local conditions.

Having computed the solution gas-liquid ratio $R_{sl}$, we can compute the liquid density based on the material balance:

\begin{equation}\label{eq:liq_density}
\rho_{liq} = \frac{\rho_{\bar{g}} R_{sl} +  \rho_{\bar{liq}}}{B_l}
\end{equation}
where $\rho_{g}$ denotes the gas density at local conditions, $\rho_{\bar{g}}$ - the water density at standard conditions, $B_g$ - the gas formation volume factor at local conditions.

The described properties are used in computing mass and momentum balances for the multiphase flow mixture for given pressure and temperature conditions along the pipe. 

\textbf{Important note on densities and viscosities computation}. As discussed above, the densities are computed based on the material balances. In OLGA, the software that simulates the true plant, the densities are computed differently.

In addition to the densities, we compute the mixture viscosity based on the volumetric balance, assuming homogeneous mixing between the phases. In OLGA, however, there are advanced options for computing the emulsion viscosities which are not based on the homogeneous mixing assumption. This influences the viscosity values which results in deviations of the friction losses values.

In this work, these simplifications are done intentionally in order to introduce additional error into computations which is then minimized by Bayesian Neural Networks and to mimic the real case, where the "true" model does not exist. To get a more accurate multiphase flow model, one may consider to compute densities and viscosities differently.

\subsubsection{Mass and momentum balance formulation and discretization}
There are two main approaches for modeling multiphase flows in pipes (\cite{nydal2012dynamic}): a two-fluid (or multi-fluid) model and drift-flux model. The two-fluid formulation is a more complex approach and considers mass and momentum equations for each fluid field such as gas, oil and water separately. In addition, liquid droplets can also be considered as a separate fluid medium. Two leading commercial multiphase flow simulators in the oil and gas industry, OLGA and LedaFlow, use this approach (\cite{bendiksen1991dynamic}, \cite{goldszal2007ledaflow}).

The drift-flux formulation treats the multiphase flow fluid as a mixture which simplifies the modeling process. This results in one momentum equation for the mixture while the mass conservation equations are typically written separately for the gas and liquid fields (\cite{holmaas2011flowmanager}). In our work, we consider the drift-flux model and then tune the model parameters such that the modeling outcomes are close to the ones produced by the two-fluid model from OLGA.

\textbf{System of equations.} We consider a steady state model, as such we do not include time derivatives into the equations. However, we do consider the mass transfer from the oil phase to the gas phase because gas bubbles out from oil when pressure decreases along the pipe. The mass transfer term is adopted from \cite{andreolli2017modeling}. The resulting mass and mixture momentum balances are:

\begin{equation}\label{eq:gas_mass}
\frac{d(\alpha_g \rho_g u_g)}{dx} = -Q_{o}^{sc} \rho_g \frac{dR_{so}}{dx}
\end{equation}

\begin{equation}\label{eq:gas_liq}
\frac{d \big( (1 - \alpha_g) \rho_{liq} u_{liq} \big)}{dx} = Q_{o}^{sc} \rho_g \frac{dR_{so}}{dx}
\end{equation}

\begin{equation}\label{eq:mix_momentum}
\frac{dP}{dx} = \frac{2 \xi_{mix}\rho_{mix} u_{mix}^2}{D_{pipe}} + \rho_{mix} g sin(\psi)
\end{equation}

where $\alpha_g$ denotes the gas volume fraction, $u_g$ - the gas phase velocity, $R_{so}$ - the solution gas-oil ratio, $u_{liq}$ - the liquid phase velocity, $Q_{o}^{sc}$ - the oil flowrate at standard conditions, $P$ - the fluid pressure, $u_{mix}$ - the mixture velocity, $\xi_{mix}$ -  the friction factor coefficient computed using mixture properties, $D_{pipe}$ - the pipe diameter, $g$ - the acceleration constant, $\psi$ - the angle of the pipe relative to the horizontal plane.

\begin{figure}[!t]
	\begin{center}
		\includegraphics[width=1.0\columnwidth]{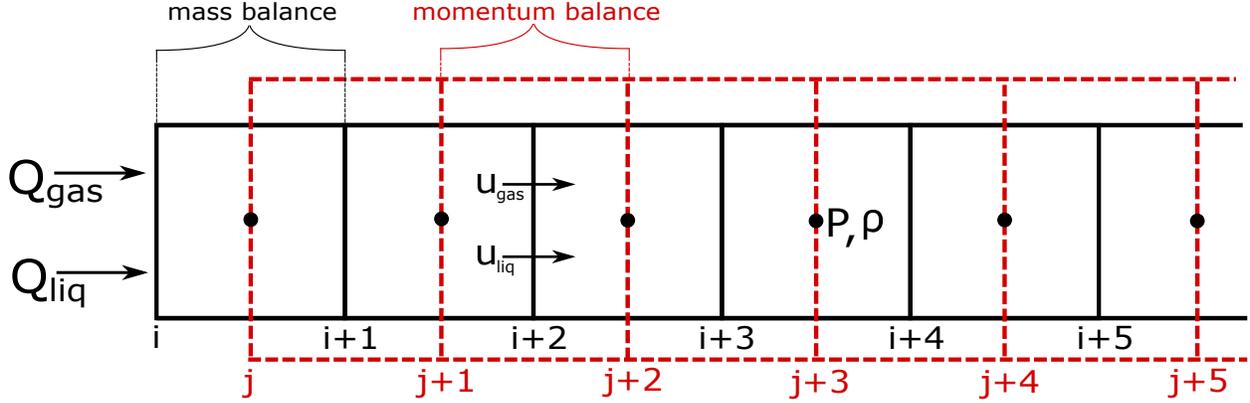}
		\caption{Staggered grid discretization scheme used to solve the system of equations of the drift-flux model} 
		\label{fig:numerical_scheme}
	\end{center}
\end{figure}

To compute the system we need the following boundary conditions:
\begin{itemize}
	\item inlet: liquid flowrate ($Q_{liq}$), inlet temperature ($T_{in}$) [in our case $T_{in}$ = $T_{out}$ because we assume isothermal flow]
	\item outlet: outlet pressure ($P_{out}$), outlet temperature ($T_{out}$)
\end{itemize}

In order to close the system of equations, the void fraction correlation is required which is used to compute the difference between the gas and liquid phase velocities and also linked with computing the local gas volume fraction (void fraction) in the pipe within the mass balance. In this work, we use a comprehensive flow pattern independent correlation by \cite{bhagwat2014flow} which shows an accurate performance for various conditions, flow regimes and fluid types. However, the model is not restricted to application of this correlation and other correlations can be used in search of a more accurate solution. Since in this paper our goal is to get a robust and physically feasible solution of the model and adjust it to the plant data, the solution accuracy is not essential. As such, it is out of the scope of this paper to compare the performance of this correlation with other ones.   

In order to compute the friction factor, we use the Colebrook equation with the mixture properties (\cite{colebrook1939correspondence}):

\begin{equation}\label{eq:mix_colebrook}
\frac{1}{\sqrt{\xi_{mix}}} = -4 log_{10} \bigg( \frac{1.256}{Re_{mix} \sqrt{\xi_{mix}}} +\frac{\epsilon}{3.7 D_h} \bigg)
\end{equation}
where $Re_{mix}$ denotes the mixture Reynolds number $\epsilon$ - the hydraulic roughness, $D_h$ - the hydraulic diameter (in our case $D_h$ = $D_{pipe}$ ).

The Reynolds number in Eq.~\ref{eq:mix_colebrook} is computed as follows:

\begin{equation}\label{eq:reynolds}
Re_{mix} = \frac{\rho_{mix} u_{mix} D_{pipe}}{\mu_{mix}}
\end{equation}

\subsubsection{Discretization and numerical scheme.} Equations \ref{eq:gas_mass}, \ref{eq:gas_liq}, \ref{eq:mix_momentum} are discretized using the control volume approach in a mesh with staggered grid. The discretization scheme is shown in Figure~\ref{fig:numerical_scheme}. The mass balances are solved within the black control volumes while the momentum balances are solved in control volumes shifted from the mass control volumes by half a mass control volume cell (shown in red). The scalar variables such as pressure, density, etc are computed in the middle of the control volumes (or at the faces of the momentum control volumes) while the velocity components computed at the at the mass control volume faces. This helps to avoid undesirable oscillating behavior of the solution of the pressure field as well as gives an opportunity to evaluate the velocities at the faces where they needed to compute advection terms of the equations, for example, $\alpha_g \rho_g u_g$ term (\cite{yang2010numerical}).



\subsection{Model solving procedure}
To solve Equations \ref{eq:gas_mass}, \ref{eq:gas_liq}, \ref{eq:mix_momentum}  within the staggered grid framework, we use Semi-Implicit Method for Pressure-Linked Equations (SIMPLE) with the first order upwind scheme for the advection terms of the mass conservation equations (\cite{wang2016numerical}, \cite{spesivtsev2013comparison}). The general solution procedure is as follows:

\begin{enumerate}
	\item Initialize the pressure and temperature fields.
	\item Compute required fluid properties at the initialized conditions.
	\item Solve the mass balance equations.
	\item Compute the velocity field.
	\item Solve the momentum balance equation.
	\item Compute the pressure field.
	\item Compute the error between the newly computed pressure field and the guessed pressures.
	\item If the pressure error is acceptable, solution is converged, if not - assign the new pressure field as the guessed pressure field (possibly with some relaxation factor) and start from Step 1.
\end{enumerate}

\section{Case study setup}

\subsection{Using Bayesian Neural Networks for tuning first principles model parameter to data}

\begin{figure}[!t]
	\begin{center}
		\includegraphics[width=1.0\columnwidth]{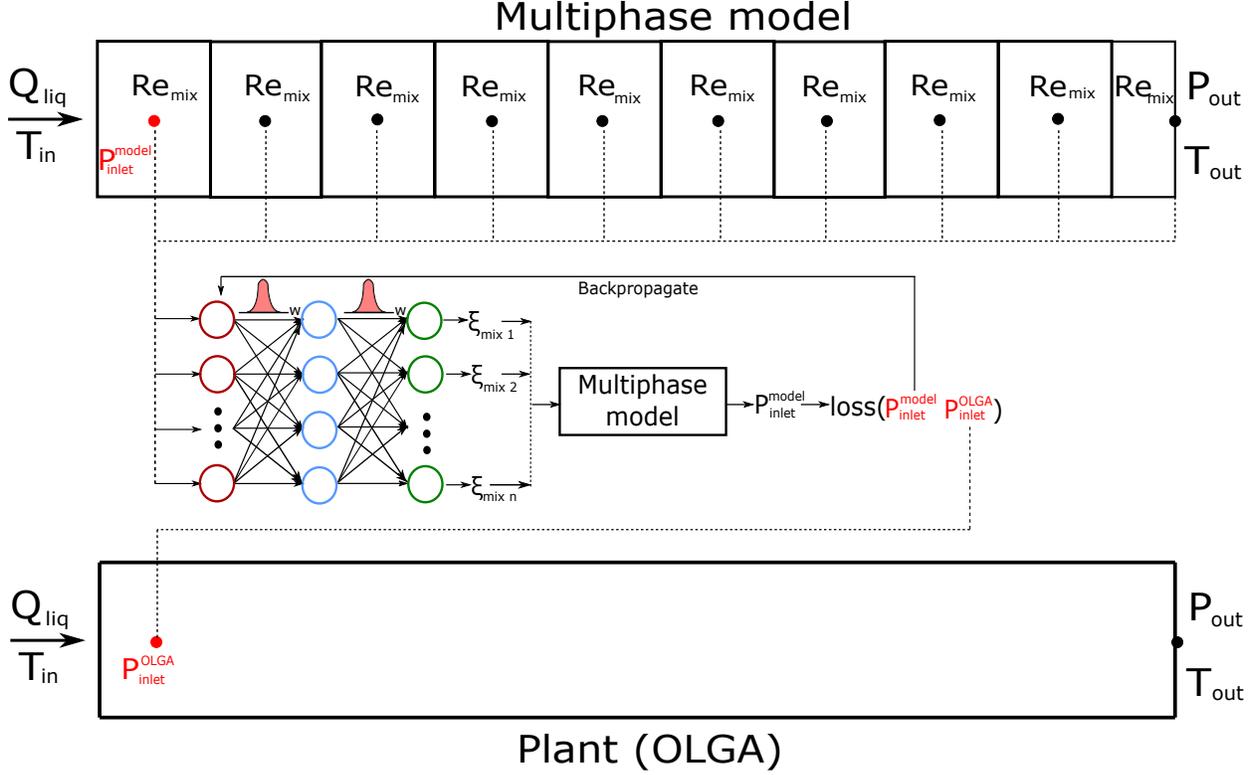}
		\caption{Training a Bayesian Neural Network for pressure drop tuning of multiphase flow in a pipe using measured inlet and outlet conditions} 
		\label{fig:case_study_train}
	\end{center}
\end{figure}

The training part of the tuning concept adaptation for the considered problem is shown in Figure~\ref{fig:case_study_train}. We selected the friction factor $\xi_{mix}$ as the parameter to be tuned. To tune the friction factor, we divide the pipe into 10 control volumes and take the calculated $Re_{mix}$ as features from each control volume. We selected Reynolds numbers $Re_{mix}$ for the features because the friction factor $\xi_{mix}$ was selected as a tunable parameter. From Eq.~\ref{eq:mix_colebrook} we see that the friction factor is a function of $Re_{mix}$, as such we will try to find almost the direct correction mapping for the friction factor $\xi_{mix}$.

The computed friction factors from the BNNs are then inserted back to the first principles model to compute the pressure at the inlet of the pipe. The computed pressure is compared with the plant measurement produced by OLGA, and the neural networks weights and biases distributions are adjusted via backpropagation.

The reason why inlet pressure is selected to be the target variable used for BNNs tuning is the fact that in oil and gas production systems inlet and outlet pressures are typically measured using sensors, such that we can use it for first principles model adjustment. In our case, we use the outlet pressure as the boundary condition of the multiphase first principles model, while the inlet pressure is used as the tuning reference. 

\begin{figure}[!t]
	\begin{center}
		\includegraphics[width=1.0\columnwidth]{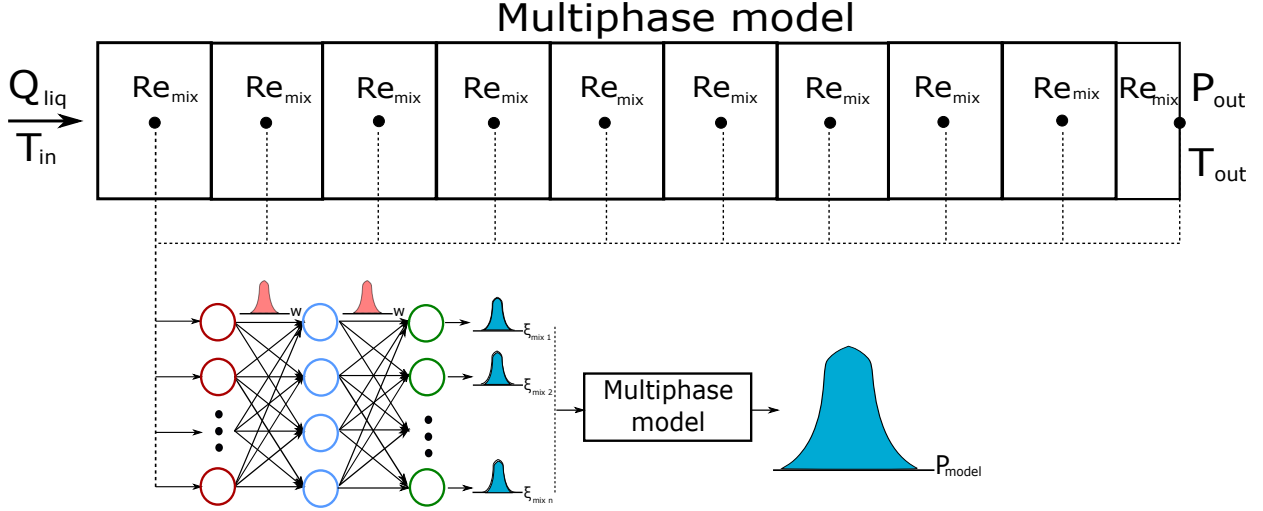}
		\caption{Prediction stage of Bayesian Neural Networks for predicting the mean and variance of the modeled variable} 
		\label{fig:case_study_test}
	\end{center}
\end{figure}

Figure~\ref{fig:case_study_test} shows the inference stage of the trained BNN at the prediction (test) time. Here, the computed $Re_{mix}$ at test conditions are inserted into the trained BNN, which produces distributions of the friction factors. Based on these distributions, the distribution of the inlet pressure is computed. 

\subsection{Case studies}

\textbf{Training set.} One of the main goals of this work is to see if BNNs are able to estimate uncertainty correctly under different process conditions. To do that, we control the distribution of the conditions in the training and test sets such that we know when the test set is within the distribution of the training set and when it is not. For the process conditions to be changed, we selected the inlet liquid flowrates. We assume that there are 1440 training points available to us. These points are split unequally such as 90\% correspond to high flowrate range values (between 0.15 and 0.25 $m^3/s$) and 10\% correspond to low flowrate values (between 0.05 and 0.15 $m^3/s$). This is done in order to see if the hybrid model with Bayesian Neural Networks produce different level of uncertainty for different flowrate ranges in the test set depending on the training set size of each range.

\begin{table}[!t]
	\begin{center}
		\begin{tabular}{ccc}
			\centering
			Parameter & Size in \% [value range] & Number of samples  \\
			\hline 
			\multicolumn{3}{c}{Training set}\\
			\hline
			\multirow{2}{*}{Inlet liquid flowrate ($Q_{liq}$)} & 10\% - [0.05 - 0.15] $m^3/s$ & 144 \\
			& 90\% - [0.15 - 0.25] $m^3/s$ &  1296 \\ 
			\hline 
			\multicolumn{3}{c}{Test set, Case 1}\\
			\hline
			\multirow{2}{*}{Inlet liquid flowrate ($Q_{liq}$)} & 50\% - [0.05 - 0.15] $m^3/s$ & 25 \\
			& 50\% - [0.15 - 0.25] $m^3/s$ & 25 \\ 
			\hline 
			\multicolumn{3}{c}{Test set, Case 2}\\
			\hline
			\multirow{2}{*}{Inlet liquid flowrate ($Q_{liq}$)} & 50\% - [0.05 - 0.15] $m^3/s$ & 25\\
			& 50\% - [0.25 - 0.30] $m^3/s$ & 25 \\ 			
		\end{tabular}
	\end{center}
	\caption{Process conditions in the training and test sets}
	\label{test_conditions}
\end{table}

\textbf{Test case studies.} We consider two case studies and each case study is performed with two types of Bayesian Neural Network: trained by MC Dropout and by Bayes by Backprop. Table~\ref{test_conditions} shows the conditions for each case study together with the training set flowrate ranges. 

In \textbf{Case 1}, we \emph{consider the same conditions} in the training and test sets. We split the data of flowrate range equally between the ranges which are used in training: 50\% for $Q_{liq}$ = \emph{[0.05-0.15]} $m^3/s$ and 50\% for $Q_{liq}$ = \emph{[0.15-0.25]} $m^3/s$. The values for training are taken from the uniform distribution within the specified ranges. 

We test the model for 50 test points, 25 points per flowrate range (high and low). The values of the flowrates are taken randomly from the specified ranges. For each value the simulation is run 5 times. It is done in this way because it produces better visualization of the results while does not deteriorate any conclusions from it.

The idea of this case study is to test capabilities of BNNs to quantify uncertainty based on the amount of information in the training set. Since we have only 10\% of the training set in the range of $[0.05-0.15] m^3/s$, our hypothesis is that the BNN will give a higher uncertainty when predicting the target variable. It is also interesting to see how in general a BNN behaves under the conditions which are within the training range.

In \textbf{Case 2}, we \emph{increase the high flowrate range from 0.25 up to 0.3} $m^3/s$ in the test set data. We introduce the new range to the model, but we do not re-train it to update the BNNs parameters. This is done to see what is the bias and uncertainty levels of the predictions for these new conditions. Table~\ref{test_conditions} shows all the values important for the case studies to consider.

\section{Results and Discussion}
\label{Results}

\subsection{Case 1 results and discussion}
In Case 1, the range of the inlet flowrates was the same as in the training set and the main idea behind this case was to see how the hybrid models based on BNNs quantify the uncertainty of the inlet pressure for high and low flowrate values with different number of training points for each flowrate range.

\begin{figure}[!t]
	\begin{center}
		\includegraphics[width=1.0\columnwidth]{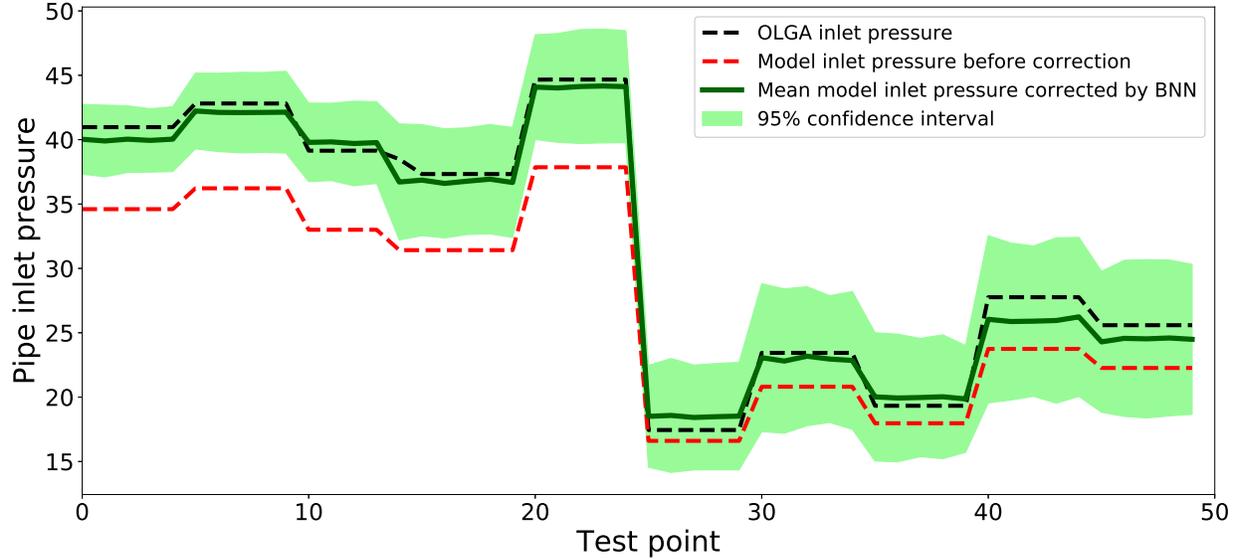}
		\caption{Estimated mean and 95\% confidence interval of the corrected multiphase flow model using MC Dropout Bayesian Neural Network for Case 1.} 
		\label{fig:MC_Case_1}
	\end{center}
\end{figure}

\begin{figure}[!t]
	\begin{center}
		\includegraphics[width=1.0\columnwidth]{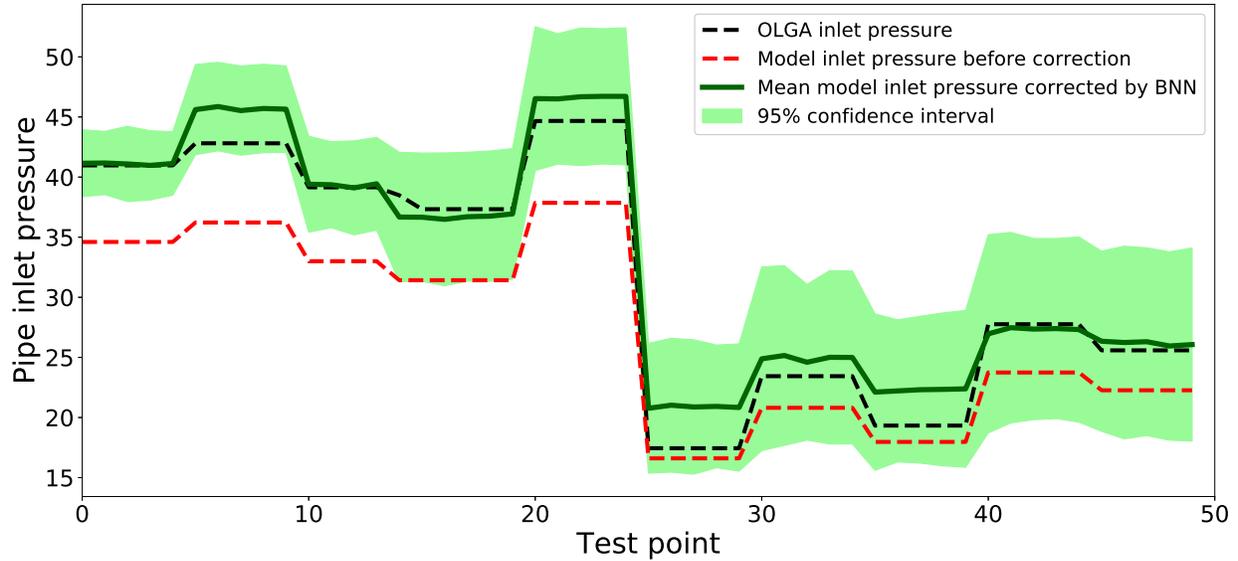}
		\caption{Estimated mean and 95\% confidence interval of the corrected multiphase flow model using Bayes by Backprop Bayesian Neural Network Case 1.} 
		\label{fig:VI_Case_1}
	\end{center}
\end{figure}

\textbf{Mean and uncertainty estimates.} Fig.~\ref{fig:MC_Case_1} and Fig.~\ref{fig:VI_Case_1} show the results of the first principles model tuning process by BNNs in Case 1. We see that both BNN types are able to tune the first principles model such that the mean estimates of the inlet pressure produced by the hybrid (first principles + BNN) models are close to the plant (OLGA) pressures. This holds for both high inlet flowrates (high pressure values) and low inlet flowrates (low pressure values). However, the Bayes by Backprop BNN shows slightly worse performance in terms of the mean estimates on the low pressure values where the number of data points is small. Due to the small number of training points in that region, the guessed priors may not have been accurate enough to obtain good posterior estimates, so that it was difficult to tune the model in that region and in some cases the tuned model works worse than the untuned version. Table~\ref{Case_1_errors} shows the mean absolute percentage errors (MAPEs), computed using Eq.~\ref{eq:MAPE}, produced by the untuned and tuned models for all flowrate ranges in Case 1. From the table we see that the errors within the high flowrate range values are smaller than for the low flowrate range for both BNNs.

\begin{equation}\label{eq:MAPE}
MAPE = \frac{1}{N} \sum_{n=1}^{N}\biggl|\frac{y_{plant} -\hat{y}_{plant}}{y_{plant}}\biggr| 
\end{equation}

where $N$ denotes the number of test points, $y_{plant}$ - the plant values of the measurement (in our case - OLGA inlet pressure), $\hat{y}_{plant}$ - the estimated mean of the plant value by hybrid (first principles + BNN) or the untuned first principles model.

\begin{table}[!t]
	\begin{center}
		\begin{tabular}{c|c|c|c}
			\multirow{2}{*}{Model} & \multicolumn{3}{c}{MAPE error}\tabularnewline
			& High flowrates & Low flowrates & Entire set\tabularnewline
			\hline 
			Untuned model & 15.65\% & 10.10\% & 12.88 \% \tabularnewline
			&&& \tabularnewline
			\begin{tabular}{@{}c@{}} Model tuned with \\ MC Dropout \end{tabular} & 1.79\% & 4.39\% & 3.09\% \tabularnewline
			&&& \tabularnewline
			\begin{tabular}{@{}c@{}} Model tuned with \\ Bayes by Backprop \end{tabular} & 2.81\% & 8.84\% & 5.83\% \tabularnewline
		\end{tabular}
	\end{center}
	\caption{Mean absolute percentage errors between the model mean outcomes and the plant pressure values in Case 1}
	\label{Case_1_errors}
\end{table}

\begin{table}[!t]
	\begin{center}
		\begin{tabular}{c|c|c|c}
			\multirow{2}{*}{Model} & \multicolumn{3}{c}{95\% confidence interval [in bar]}\tabularnewline
			& High flowrates & Low flowrates & Entire set\tabularnewline
			\hline 
			\begin{tabular}{@{}c@{}} Model tuned with \\ MC Dropout \end{tabular} & 3.52 & 5.27 & 4.40 \tabularnewline
			&&& \tabularnewline
			\begin{tabular}{@{}c@{}} Model tuned with \\ Bayes by Backprop \end{tabular} & 4.26 & 6.95 & 5.61 \tabularnewline
		\end{tabular}
	\end{center}
	\caption{95\% confidence intervals produced by hybrid Bayesian models in Case 1}
	\label{Case_1_conf_intervals}
\end{table}

As for the uncertainty estimates level, in Fig.~\ref{fig:MC_Case_1} and Fig.~\ref{fig:VI_Case_1} we observe that the average uncertainty estimates for the low pressures are higher, than for the high pressure values. The values of the 95\% confidence intervals are shown in Table~\ref{Case_1_conf_intervals}. We see that the estimates of the confidence intervals produced by the different types of BNNs are quite similar, however, the BNN trained Bayes by Backprop produce higher uncertainty levels.

\subsection{Discussion of Case 1 results} Our observations of the simulation results have been hypothesized prior to the training and confirmed at the results stage. For instance, the results confirmed that both BNN types are able to correctly qualitatively identify the uncertainty levels of the first principles models parameters (friction factors) based on the training size distributions. The distribution of the friction factors have been then introduced into the first principles models which resulted in the uncertainty level estimates of the inlet pressure values. The results also show that in the regions where BNNs have less prior information, they have less accurate interpolating capabilities, which resulted in less accurate mean estimates and larger uncertainty. On the other hand, we see that in most of the cases even a small training set was enough for the neural networks to adjust the first principles models to the plant conditions. However, this is mainly caused by the fact that no noise was introduced into the data. This has been done intentionally to prove the concept of tuning parameters by BNNs, rather than additionally challenge them to fit the data at hand. In general, more noisy data would result in the need of more tuning data.

\textbf{Training difficulties of the Bayes by Backprop algorithm.} In general, the success of neural network training depends on the parameters (weights and biases) initialization. This is especially important for Bayesian Neural Networks because the poor prior distribution of the parameters will usually result in poor posteriors, which in turn results in inaccurate and uncertain predictions.

Since the training procedure of the MC Dropout network is similar to conventional training, in this work we used the initialization procedure by \cite{he2015delving} which has been proven as a robust initialization technique. As for the Bayes by Backprop, we used the normally distributed priors of the neural network weights as suggested in the original paper by \cite{blundell2015weight}. However, in practice we observed that guessing a good prior is difficult and we met problems trying to find a suitable prior which results in a robust tuning, meaning that the guess prior had been leading to bad posteriors and poor simulation results.

To overcome the obstacles with tuning the Bayes by Backprop BNN, we initialized the priors close to the weights distributions produced by the tuned MC Dropout BNNs. This resulted in a relatively robust tuning procedure. However, to make it even more robust, we performed 3 samples of the weights per epoch to compute the approximated cost shown in Eq.~\ref{eq:KL_cost_approx}, which resulted in a less stochastic behavior of the cost function and more robust tuning. Therefore, as a result of our investigations, we suggest a similar approach for guessing a good initial prior when performing first principles model tuning using the Bayes by Backprop algorithm.

One could argue that there is no point in creating a hybrid (first principles + BNN) model using the Bayes by Backprop algorithm if the MC Dropout is more robust in tuning. However, this may not always be the case and in the literature there are various examples when MC Dropout underestimated the uncertainties if compared to the Bayes by Backprop algorithm (\cite{blundell2015weight}). In fact, this is what we also observe in our case. Despite that the estimated uncertainty levels for both BNNs types are quite similar, the levels produced by Bayes by Backprop are larger (see Table~\ref{Case_1_conf_intervals}) and in fact may be more accurate. As such, when the level of the hybrid model uncertainty is critical for the problem at hand, for instance for the subsequent use in robust optimization of the process, one may consider using the Bayes by Backprop algorithm rather than MC Dropout.

\textbf{Data noise and overfitting.} One more important fact to discuss here is that we assumed that the plant measurements do not have noise which made it easier for the BNNs to tune the first principles model to the plant conditions. In real operation, this is typically not the case. This will most likely result in the fact that more measurement data will be required to tune the models accurately. Moreover, the tuning itself will be more difficult because data noise may cause overfitting. At the same time, a strong feature of Bayesian Neural Networks is the fact that they are less prone to overfitting than conventional maximum likelihood neural networks due to model parameters averaging. In addition, the data noise can be learned from the data. This can be done via selecting the noise level as the learnable BNN's parameter. This will result in the fact that the noise will be accounted when making mean and uncertainty estimates of the target variable.

\subsection{Case 2 results and discussion}
In Case 2, the range of the inlet flowrate was not the same as in the training set for the high flowrate range. As such, the main idea behind this case was to see how BNNs estimates the model bias and uncertainty when the plant process conditions change. In this case, the inlet flowrate is outside the training set, however, since we use the Reynolds numbers as the features, the feature distribution is not necessarily fully outside the training set.

\begin{figure}[!t]
	\begin{center}
		\includegraphics[width=1.0\columnwidth]{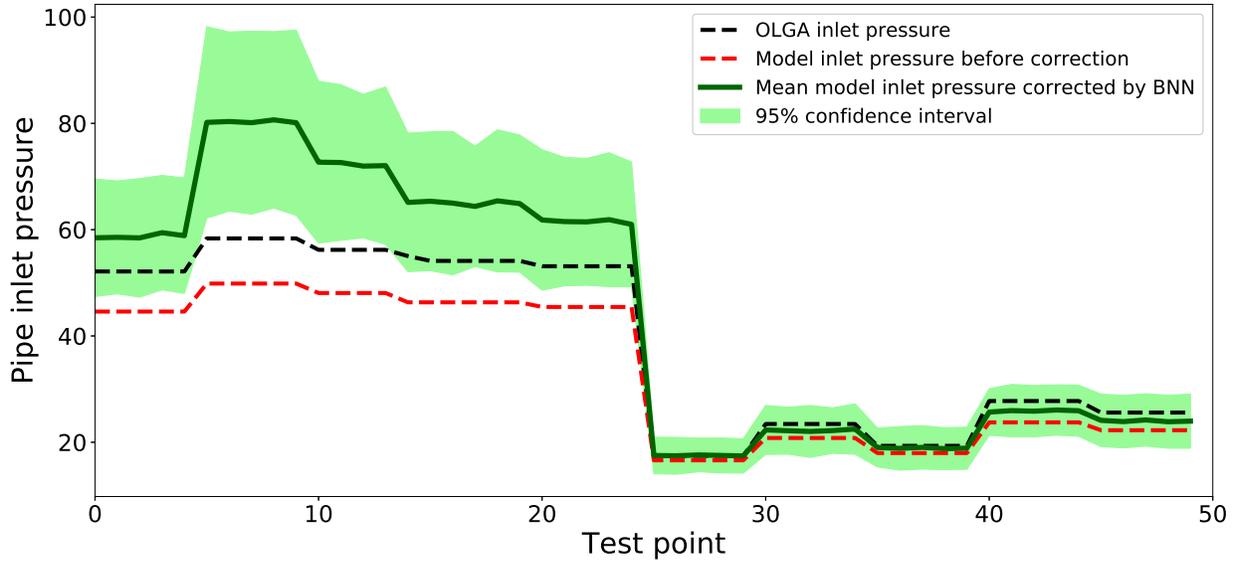}
		\caption{Estimated mean and 95\% confidence interval of the corrected multiphase flow model using MC Dropout Bayesian Neural Network for Case 2.} 
		\label{fig:MC_Case_2}
	\end{center}
\end{figure}

\begin{figure}[!t]
	\begin{center}
		\includegraphics[width=1.0\columnwidth]{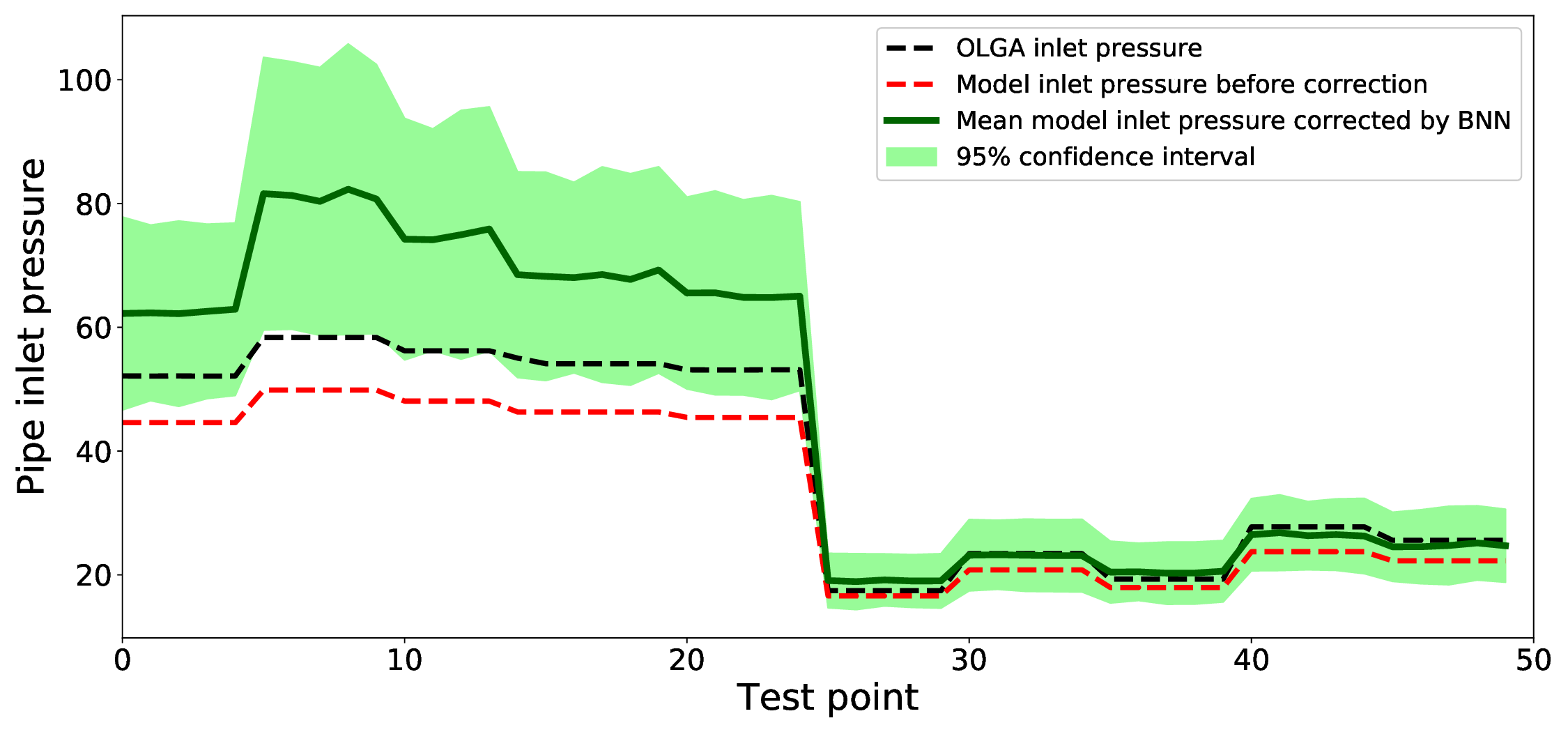}
		\caption{Estimated mean and 95\% confidence interval of the corrected multiphase flow model using Bayes by Backprop Bayesian Neural Network Case 2.} 
		\label{fig:VI_Case_2}
	\end{center}
\end{figure}

\textbf{Mean and uncertainty estimates.} Fig.~\ref{fig:MC_Case_2}, Fig.~\ref{fig:VI_Case_2}, Table~\ref{Case_2_errors} and Table~\ref{Case_2_conf_intervals} show the results of the first principles model tuning process by BNNs in Case 2. From the figures and tables we see that where the process conditions do not change (low pressure values), the uncertainty level and the mean predictions error are low and in fact the same as in the Case 1 because the conditions are the same. This is not the same as for the high flowrate case. We see that in addition to the bias, the uncertainty level of predictions are several times larger than for the low flowrate cases and even worse than for the untuned model. This is true for both BNNs types.

Such results have been expected since the flowrate values which correspond to the high pressure values in Case 2 are outside the training range. This resulted to the fact that the difference between the mean estimates and the actuate plant values are high. But what is more important, we can see that the both BNNs types are able to correctly estimate the qualitative uncertainty levels depending on how far the test data points are from the training set, as we see that for the highest pressure value is uncertainty is the largest for both BNNs types. We discuss the usefulness and practical applicability of the obtained results in the next subsection.

\begin{table}[!t]
	\begin{center}
		\begin{tabular}{c|c|c|c}
			\multirow{2}{*}{Model} & \multicolumn{3}{c}{MAPE error}\tabularnewline
			& High flowrates & Low flowrates & Entire set\tabularnewline
			\hline 
			Untuned model & 15.65\% & 10.10\% & 12.88 \% \tabularnewline
			&&& \tabularnewline
			\begin{tabular}{@{}c@{}} Model tuned with \\ MC Dropout \end{tabular} & 22.58\% & 4.39\% & 13.48\% \tabularnewline
			&&& \tabularnewline
			\begin{tabular}{@{}c@{}} Model tuned with \\ Bayes by Backprop \end{tabular} & 27.88\% & 8.84\% & 18.36\% \tabularnewline
		\end{tabular}
	\end{center}
	\caption{Mean absolute percentage errors between the model outcomes and the plant pressure values in Case 2}
	\label{Case_2_errors}
\end{table}

\begin{table}[!t]
	\begin{center}
		\begin{tabular}{c|c|c|c}
			\multirow{2}{*}{Model} & \multicolumn{3}{c}{95\% confidence interval [in bar]}\tabularnewline
			& High flowrates & Low flowrates & Entire set\tabularnewline
			\hline 
			\begin{tabular}{@{}c@{}} Model tuned with \\ MC Dropout \end{tabular} & 13.49 & 5.27 & 9.38 \tabularnewline
			&&& \tabularnewline
			\begin{tabular}{@{}c@{}} Model tuned with \\ Bayes by Backprop \end{tabular} & 17.6 & 6.95 & 12.28 \tabularnewline
		\end{tabular}
	\end{center}
	\caption{95\% confidence intervals produced by hybrid Bayesian models in Case 2}
	\label{Case_2_conf_intervals}
\end{table}

\subsection{Discussion of Case 2 results}

\textbf{Importance of uncertainty levels for monitoring of process system plants.} 

\emph{Process measurement drift / Non-observable process conditions change}

The obtained results lead to important conclusions of using uncertainty-based hybrid machine learning models from the process operations point of view. In many cases, the process measurements drift over time which is hard to identify in practice. Let us assume that the inlet pressure drifts over time and we are not aware of this situation. Having uncertainty estimates about our predictions will help us to identify the measurement drift. This is because, \emph{assuming that we know the inlet flow conditions}, we know that the flow conditions have not change and for any particular point in time we get high confidence in our predictions (low uncertainty levels). At the same time, we see that the inlet pressure estimates deviates from the measurements. This will be a most likely caused by the fact that the process measurement drifted over time. 

In addition, we may be able to identify other process condition changes. For instance, again, assuming that we know or measure inlet flowrates, we may see the deviation between the mean estimates of the proposed model and measurements. Knowing that the observed conditions have not changed, there is a possibility that non-observable process conditions change. In the multiphase flow pipeline, this can be for instance, wax and hydrate depositions which increased the friction loss and caused additional pressure drop. As such, hazardous situation can be avoided using the proposed methods with a much higher confidence that using just untuned non-bayesian first principles models.

\emph{The need for model recalibration}

There are less hazardous situations when Bayesian hybrid models can be effectively used. This is when we know that the observable process conditions changed over time and the bias and uncertainty levels of the models increased. This is the exact situation which we have provided in our work. We changed the inlet flowrate and observed that the difference between the referenced pressures and produced mean estimates became high for the high flowrate values and this difference has been also confirmed by the high uncertainty levels. This is a clear situation when the model needs recalibration, because the deviations and high uncertainties are caused by the fact that the model has not seen such process conditions before because the plant has ot been operated around this point. After recalibration the model can be used successfully again for the new conditions.

\section{Conclusions}
\label{Conclusions}
In this paper, we propose the general framework of creating hybrid machine learning models by tuning first principles models with Bayesian Neural Networks and show applications of such a framework based on tuning a steady state three-phase multiphase flow model. For the Bayesian Neural Networks training, we used variational approximation methods such as MC Dropout and Bayes by Backprop. The multiphase flow model consists of two main parts: fluid properties model represented by the Black Oil approach and the hydrodynamic drift-flux model solved by the SIMPLE numerical integration scheme over the staggered numerical grid.

We found that by tuning the first principles model using Bayesian Neural Networks, it is possible to adjust them to the previously observed process conditions such that the tuned model correctly represents the process. As such, the Bayesian Neural Networks were found to be a good tool for tuning the first principles models despite a relatively small dataset size. The main advantage of using the Bayesian Neural Networks was found to be the ability to correctly estimate the uncertainty level of predictions depending on the number of available training data points, such that for a smaller part of the training distributions, the produced levels of uncertainties were higher. We also found that Bayesian Neural Networks are able to correctly identify the level of uncertainty when the process conditions become outside the training range.

In addition, we discussed how the proposed method can be used to identify the measurement drift in a plant, change of unobservable process conditions or when the model recalibration is needed. We found that the proposed methods can be of a great importance when making decisions during monitoring of process engineering systems.

As for the difference between the Bayesian Neural Networks, we found that the ones trained with the Bayes by Backprop algorithm produce larger uncertainty levels, slightly less accurate in terms of the mean estimates and are much harder to fit than the BNNs trained with the MC Dropout algorithm. As an ad-hoc training approach, we proposed to train an MC Dropout BNN first, sample the weights and use it as a prior to the weights for the Bayes by Backprop training.

In general, we believe that the first principles models tuned by Bayesian Neural Networks will become a great, robust and highly usable tool in the near future in process conditions monitoring.  This is because the computational power increases over time which allows training deep Bayesian Neural Networks for any process at hand. In addition, such an approach will provide more degrees of freedom when making correct decisions to either change the measurement sensors, perform additional maintenance check and operation, for instance, pigging, or perform recalibration of Digital Twins models which represent a real time process plant behavior.

\section*{Acknowledgments}
The authors gratefully acknowledge the financial support from the center for research-based innovation SUBPRO, which is financed by the Research Council of Norway, major industry partners, and NTNU.



\bibliographystyle{elsarticle-harv} 
\bibliography{paper}


\end{document}